# Algorithmic Frameworks for the Detection of High-Density Anomalies

Ralph Foorthuis
Digital & Technology, HEINEKEN
Amsterdam, The Netherlands
ralph.foorthuis@heineken.com

*Abstract* — This study explores the concept of high-density anomalies. As opposed to the traditional concept of anomalies as isolated occurrences, high-density anomalies are deviant cases positioned in the most normal regions of the data space. Such anomalies are relevant for various practical use cases, such as misbehavior detection and data quality analysis. Effective methods for identifying them are particularly important when analyzing very large or noisy sets, for which traditional anomaly detection algorithms will return many false positives. In order to be able to identify high-density anomalies, this study introduces several non-parametric algorithmic frameworks for unsupervised detection. These frameworks are able to leverage existing underlying anomaly detection algorithms and offer different solutions for the balancing problem inherent in this detection task. The frameworks are evaluated with both synthetic and real-world datasets, and are compared with existing baseline algorithms for detecting traditional anomalies. The Iterative Partial Push (IPP) framework proves to yield the best detection results.

## I. INTRODUCTION

Academia and practice are dependent on increasingly larger collections of data. These collections are mined to obtain valuable insights, for example by creating descriptive overviews, training predictive models, and detecting clusters, associations and anomalies [1, 2, 3]. Anomaly detection (AD) is the task of analyzing the data to detect unusual occurrences. An anomaly is a case, or a group of cases, that is in some way different from the rest and does not fit the general patterns in the dataset [4, 5, 6, 38]. Such cases are often also referred to as outliers, deviants, novelties or discords. Anomaly detection can be used for a wide variety of purposes, such as fraud detection [1, 9, 11], security and process monitoring [9, 10], data quality analysis [17] and data preparation prior to statistical modelling [1, 5, 8].

Cases are traditionally considered more anomalous if they deviate w.r.t. more attributes. AD algorithms therefore generally assign more extreme anomaly scores to cases that have larger numbers of deviating attribute values [12]. However, it has been acknowledged that anomalies can be deviant with regard to a single attribute or a certain subspace (i.e. a subset of attributes), while exhibiting normal behavior w.r.t. the other attributes [1, 13, 14]. This is relevant for practice, because there is significant value in identifying the cases that are not the most anomalous on all accounts, but that exhibit both normal and abnormal characteristics. Detecting such cases is particularly relevant for unmasking misbehavior. People or organizations attempting to commit fraud will try to keep their actions as close as possible to normal, legitimate actions [3, 10, 15]. For this reason tax returns with many deviant values are often considered to be less risky than cases with only one or two unusual attribute values [12]. In a similar vein, unusual banking transactions that exhibit many normal properties, such as common sums of money, may point to fraudulent behavior [11]. In data quality analysis it is also relevant to detect cases that are both normal and abnormal, i.e. are anomalous and yet located in the normal (i.e. high-density) regions of the data. This may typically inspire the implementation of strict quality verification rules or structural software improvements [16, 17].

This study therefore explores the concept and detection of high-density anomalies (HDAs). These are occurrences that deviate from the norm but in some subspace are located in relatively high-density regions, i.e. are positioned amongst or are a member of the most normal cases. Semantically they can be interpreted as deviant occurrences that hide in normality. Contrary to traditional anomalies, which are typically conceptualized as low-density (isolated) cases [18], high-density anomalies hide between the most normal data points. As such they are not detected by traditional AD algorithms or, if identified, get assigned a modest anomaly score. To the best of my knowledge the concept of high-density anomalies was introduced in [19], which showed that the method of discretization impacts the kinds of anomalies that are detected. No other published work seems to investigate the concept of HDAs or approaches to identify them, nor are datasets with explicitly labeled HDAs publicly available. This study therefore introduces and evaluates the IPP and HMDH approaches for detecting high-density anomalies and compares their performance with existing non-HDA approaches. To explain and illustrate the concept of HDAs ample data plots are presented. In addition, datasets with labeled HDAs as well as R code have been published online.

This paper proceeds as follows. Section II presents background theory and related research. Section III introduces the new approaches, which are evaluated in Section IV. Sections V and VI present the discussion and conclusion.

## II. THEORY

For a proper understanding of the concept of HDAs it is valuable to first discuss the various types of anomalies that may reside in datasets. The typology of anomalies presented in [6, 38] offers a theoretical underpinning of the nature of different anomaly types and uses two main dimensions:

- *Types of Data*: The data types of the attributes that are involved in the anomalous character of a deviant case. These can be *quantitative* (numerical, e.g. volume or length), *qualitative* (categorical, code- or class-based, e.g. gender or animal species) or *mixed* (when both types are involved).

- *Cardinality of Relationship*: The way in which the various attributes relate to each other when exhibiting anomalous behavior. If no relationship between the attributes (variables) exists to which the anomalous character of the deviant case can be traced back, the relationship is said to be *univariate* and the analysis can assume independence between them. On the other hand, if the deviant behavior of the anomaly lies in the relations between its variables, i.e. in the combination of its attribute values, then the relationship is said to be *multivariate*. This means the attributes need to be described and analyzed jointly, not separately, in order to account for the relationships between them.

| | | Types of Data | | |
|---|---|---|---|---|
| | | Quantitative attributes | Qualitative attributes | Mixed attributes |
| **Cardinality of Relationship** | Univariate | Type I<br>Uncommon number anomaly | Type II<br>Uncommon class anomaly | Type III<br>Simple mixed data anomaly |
| | | Atomic univariate anomaly | | |
| | Multivariate | Type IV<br>Multidimensional numerical anomaly | Type V<br>Multidimensional categorical anomaly | Type VI<br>Multidimensional mixed data anomaly |
| | | Atomic multivariate anomaly | | |
| | | Type VII<br>Aggregate numerical anomaly | Type VIII<br>Aggregate categorical anomaly | Type IX<br>Aggregate mixed data anomaly |
| | | Aggregate anomaly | | |

Fig. 1. Typology of anomalies [6, 38] (only the top six types are relevant here)

These two dimensions yield six relevant types of anomalies. The focus in this study is on individual (atomic) data points in independent data, not on aggregates. The anomaly types are described below and illustrated in Fig. 2. (Note: some of this paper's visuals can be best be viewed by zooming in on a screen; qualitative attributes are represented by colors and shapes).

- *Type I – Uncommon number anomaly*: A case with an extremely high, low or otherwise rare value for one or several individual quantitative attributes. Extreme value outliers are typically considered in traditional univariate statistics. The leftmost case in Fig. 2 is an example, being extreme on a single attribute, namely *x1*.

- *Type II - Uncommon class anomaly*: A case with an uncommon class value for one or several individual qualitative variables. Such values can be few and far between (i.e. rare) or occur only once (i.e. truly unique). Fig. 2 contains three Type II anomalies with unique shapes.

- *Type III - Simple mixed data anomaly*: A case that is both a Type I and Type II anomaly, i.e. with at least one extreme value and one rare class.

- *Type IV - Multidimensional numerical anomaly*: A case that does not conform to the general patterns when the relationship between quantitative attributes is taken into account, but that does not have extreme or isolated values for any of its individual attributes. Detection therefore requires several numerical attributes that are analyzed jointly. The example in Fig. 2 is multivariately isolated, with extreme individual values for neither *x1* nor *x2*.

- *Type V - Multidimensional categorical anomaly*: A case with a rare combination of class values. Two or more qualitative attributes thus need to be analyzed jointly to discover it. Both examples in Fig. 2 have a regular individual shape and color, but a unique combination thereof.

- *Type VI - Multidimensional mixed data anomaly*: A case with a deviant relationship between its quantitative and qualitative attributes. The anomalous case generally has a categorical value or a combination of categorical values that in itself is not rare in the dataset as a whole, but is only rare in its numerical neighborhood. Type VI anomalies typically seem misplaced or mislabeled, as illustrated by the purple circle in Fig. 2 (which seems to belong to the purple cluster).

*A. High-Density Anomalies*

Not all the anomalies described and illustrated above are high-density anomalies. To assess whether a deviant occurrence is a HDA, it is necessary to acknowledge a third dimension for defining anomalies, the *data distribution* [38]. This dimension focuses on the dispersion of data points throughout the space. The bottom two Type II anomalies in Fig. 2 are HDAs because they are located in high-density regions. The Type II occurrence in the upper left quadrant, however, is a low-density anomaly as it lies in a sparse neighborhood. Likewise, the Type V anomaly in the upper right quadrant is a HDA, while the Type V case below it is a low (or moderate) density anomaly. Finally, a Type VI anomaly, per definition, lies in-cloud, i.e. amongst other cases. However, the Type VI occurrence in Fig. 2 can also be said to be positioned in a moderately (rather than highly) dense area. Note that the difference between 'high-density' and 'low-density' thus refers to a continuous rather than a binary property.

Some anomaly types (I, III and IV) are *necessarily* isolated and will never be HDAs, some anomaly types *may* be isolated or positioned within a cluster (II and V), and other types are *always* positioned within a moderately or highly dense cluster (VI). However, the concept of high-density anomalies can be extended beyond what is discussed in this section (see the Discussion for more on this).

*B. Related Research*

Previous research has acknowledged cases that have both normal and abnormal characteristics. For example, a *fringelier* is an unusual data point, albeit one that occurs more often than seldom [20]. With its position of about three standard deviations from the majority of the numerical data, such a case can be considered a modest Type I anomaly. This is closely related to the *in-disguise* anomalies acknowledged in [21], which are defined as exhibiting only a minor deviation from the normal pattern. One may indeed be interested in subtle deviations, rather than in radically different occurrences [22].

The concept of *inliers* is also relevant for the current study, although the term is used inconsistently in the literature. In [23] an inlier is defined as a case that lies in the interior of the distribution, but is in error nonetheless. An example is a temperature value that is measured in degrees Celsius but reported in degrees Fahrenheit. It may be impossible to detect such occurrences without additional information, such as data from the previous year. The case then is thus normal in a subspace of the variables, but multivariately anomalous in the broader set (i.e. is of Type IV, V or VI). In [24] a very similar definition of an inlier is used, although without it necessarily being erroneous.

The above refers to inliers as cases that are both normal and abnormal. However, note that other authors employ a different definition of inliers, namely simply as non-outliers [25, 26]. By definition, an inlier then refers to the vast majority of the cases that can simply be regarded as normal in all respects. Going one step further, the term inliers may be used for the roughly 60% of most normal and representative cases, which could be relevant if the aim is to obtain an estimate of population parameters using a smaller sample [27]. For the present study the definitions of [23, 24, 27] are the most relevant, as these focus on cases that have both normal and abnormal characteristics, or on the cases positioned in the most dense regions.

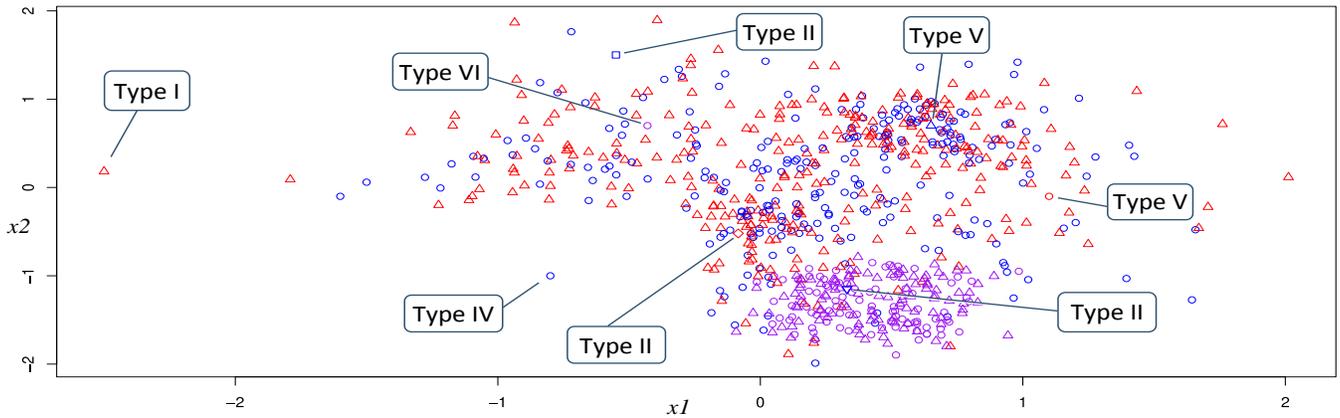

Fig. 2. Illustration of various anomaly types using two numerical (*x1* and *x2*) and two categorical (represented as color and shape) attributes.

The *singular outlier* is another related topic [12]. This refers to a case that exhibits a deviant value for only one or two attributes, while showing common behavior for all others. These outliers are typically Type I anomalies.

To conclude, there are various conceptualizations that emphasize that anomalies may have both normal and abnormal characteristics. When all relevant data is available, these conceptualizations generally boil down to a classic anomaly type known from the literature, as described in Fig. 1. Few studies explicitly emphasize that some properties of the anomalous occurrences are normal and have their algorithms deal with that directly. Only the research of [7, 19] targets anomalies in the densest areas, which is also the topic of the current study.

## III. DETECTION APPROACHES

This section describes various non-parametric algorithms and algorithmic frameworks for the unsupervised detection of (high-density) anomalies.

### A. Existing baseline algorithms

Existing general-purpose AD algorithms can be expected to identify HDAs, albeit with a considerable number of false positives due to purely isolated cases also being declared deviant. All these methods directly or indirectly yield some form or proxy of density assessment. They will be used as baseline algorithms against which the novel algorithmic frameworks can be compared, and may also be used as the underlying algorithms within these frameworks.

*Distance-based methods*: These methods calculate the global distances between data points, with the cases having the largest distances to their nearest neighbors being the most anomalous [25, 31, 32]. These methods usually rely on numerical data, meaning the data must be pre-processed first (i.e. creating dummy variables and normalizing the data). This study uses KNN-AGG [31, 32] and the scalable, sampling based QSP algorithm [25].

*Local density-based methods*: These methods not only target the global anomalies detected by distance-based methods, but also outliers that are only isolated in terms of their local neighborhood [33]. This may be relevant in sets with several clusters that have different densities. Such methods generally also require numerical data and consequently a pre-processing phase. This study uses LOF [33, 34].

*Global density-based methods*: These methods focus on e.g. frequencies or kernel-density to detect global anomalies [cf. 30]. This study uses SECODA, which allows analysis of numerical and categorical data by employing iterative discretization of quantitative attributes [7, 17]. With its standard settings this algorithm performs equiwidth discretization, i.e. equal interval binning, which divides the range of an attribute's continuous values into *b* bins of the same value interval. It also allows for equidepth discretization, i.e. equal frequency binning, which divides a continuous attribute into *b* bins that each contain the same number of cases. Research has shown that equidepth binning can directly target in-cloud and even high-density anomalies [ibid.]. The reason for this is that equidepth binning ignores numerically isolated cases, because the discretization intervals in sparsely populated areas get stretched so as to fill the bins with an equal amount of data points. Of these existing methods only SECODA with equidepth discretization can thus be expected to perform relatively well w.r.t. detecting HDAs.

### B. Proposed algorithmic frameworks

This section presents *algorithmic frameworks* for HDA detection introduced by the current study. Algorithmic frameworks leverage underlying general-purpose algorithms [cf. 28], such as those described above. The underlying algorithms yield scores on both anomalousness and neighborhood density, which then have to be optimally balanced. Let $X$ represent an $n \times p$ matrix. Let $x_{g,h}$ represent the matrix value of the $h$th attribute for the $g$th row (case), with $g = 1, 2, \ldots n$ and $h = 1, 2, \ldots p$. Let $y$ represent a column vector with $n$ rows unless it is a subset. The $\oplus$ symbol represents concatenation of different chunks and/or types of data. The function AnomalyDetectionAlgorithm() represents executing an underlying AD method such as LOF. Finally, the **# green remarks** provide explaining comments within the pseudocode. See the Remarks section for downloading the complete implementations in R.

The frameworks – IPP and HMDH – assume that the underlying algorithms analyze the joint distribution, i.e. take into account the multivariate relationships, and that the lowest scores they return represent the most anomalous cases.

*Iterative Partial Push (IPP)*: The key idea behind the IPP framework is to iteratively filter ('push') out the isolated cases from the entire set of anomalies. The pseudocode is presented below. The algorithm starts by running the underlying AD algorithm, e.g. KNN-AGG, using all attributes (yielding vector *aas*) and a second run on only the numerical attributes (yielding *ads*). The density assessment will be based on these numerical attributes ('dentributes') used for *ads*. A series of iterations will then filter out the isolated cases identified in *ads* from *aas*. The QD (QuantileDenominator) setting determines the number of iterations and the granularity of each filter run. In the first itera-tion the first quantile of *aas* (which contains *aas*' most extreme general anomalies) is cleaned up, by filtering away from it the extremely isolated cases that populate the first quantile of *ads*. In the second iteration this is performed for the second quantile, and so on. The *QFB* (QuantileFilterBoost) is used to broaden each iteration's numerical quantile proportion, so that the optimal fraction of isolated cases is filtered out. The *QFB* represents the degree of dispersion in the data distribution (relative to what can be expected by chance) and acts as the noise filter level. If the degree of dispersion is low (i.e. the degree of clustering is high), then the *QFB* will be low and only few additional cases are filtered away. However, if not much high-density clustering exists then many cases are scattered and the set is thus 'noisy', meaning more low-density cases need to be filtered out to avoid false positives. The *QFB* can be manually set or determined automatically, the latter of which uses the intrinsic density and optimal arity assessment of SECODA (albeit a simpler discretization process could also be run, e.g. with a fixed number of bins). The pseudocode is shown at the top right of this page.

As a final step in each iteration the score for the selected cases is determined. This is done by concatenating the iteration *i*, a decimal point, and the extremity of each case in the current iteration (determined after sorting the cases ascending on *aas* and descending on *ads*). For example, the score of a case that in the *third* iteration's subset of 19 cases is the *second* most extreme will get assigned a score of 3.02. After the last iteration, IPP returns a vector of gradual anomaly scores, with lower values representing more extreme HDAs. The pseudocode of IPP is:

**Algorithm:** Iterative Partial Push
**Input data:** *D*, the original matrix with *n* cases and *p* attributes
**Input parameters:** *QFB* = -9999 # QuantileFilterBoost
                       *QD* = 100 # QuantileDenominator
**Output:** *hds*, a vector of high-density anomaly scores for all cases in *D*, with $hds_g$ representing the individual score
**Key local variables:** *i*, the current iteration.
    *aas*, vector with general anomaly scores (for all anomaly types)
    *ads*, vector with density scores (based on continuous attributes)
    *qp*, the quantile proportion, with $0 < qp \leq 1$
**begin**
    $i \leftarrow 0$; *continue* ← TRUE # Set initial values
    *aas* ← AnomalyDetectionAlgorithm(*D*) # AD on full dataset
    *ads* ← AnomalyDetectionAlgorithm($D_c$) # AD on continuous attributes
    **if** *QFB* = -9999 # If not specified, then calculate QuantileFilterBoost
        *QFB* ← CalculateQFB($D_c$)
    **end if**
    **while** *continue* = TRUE **do**
        $i \leftarrow i + 1$
        $qp \leftarrow i/QD$ # Determine quantile proportion
        *aasIDs* ← set of the *aas* IDs with each case such that
            its $aas_g$ < the *qp* quantile value # Identify this iteration's most extreme anomalies
        $qp \leftarrow \min((QD-1)/QD, ((i+QD/100 \cdot QFB)/QD))$ # Determine quantile
        proportion for numerical analysis
        *adsIDs* ← set of the *ads* IDs with each case such that
            its $ads_g$ < the *qp* quantile value # Identify this iteration's most extreme anomalies
            w.r.t. numerical (continuous) variables
        *hdsIDs* ← *aasIDs* △ *adsIDs* # Discard isolated cases by determining the symmetric
            difference of *aasIDs* and *adsIDs*
        *nullIDs* ← IDs of cases in *hds* that still have a NULL value
        *g* ← *hdsIDs* ∩ *nullIDs* # Intersect to identify cases to update
        sort *g* ascending on attribute *aas* and descending on attribute *ads*
        **for each** $g_g \in hds$ **do** # Update empty *hds* scores of cases identified in *g*
            $hds_g \leftarrow i \oplus "." \oplus$ the row number of $g_g$ # Use sorted *g* to set decimals
        **end for**
        **if** $i \geq QD$
            *continue* ← FALSE # All iterations have been run
        **end if**
    **end while**
    **if** $hds_g$ = Null # Check if missing scores exist (as the isolated cases are filtered away)
        $hds_g \leftarrow 1 + \max(ads) - ads_g + QD$ # Set high score for isolated case
    **end if**
    **return** *hds* # Return full anomaly score vector as the end result
**end**

**Algorithm:** CalculateQFB # Calculate QuantileFilterBoost
**Input data:** *ads*; $D_c$, the original matrix with *n* cases and *p* continuous (numerical) attributes
**Output:** *QFB*, the degree of dispersion (noise or clustering) in the distribution
**begin**
    *p* ← the number of continuous variables in $D_c$
    *UltimateArity* ← number of SECODA iterations required to determine *ads*
    $ExpectedRandomDensity \leftarrow \frac{n}{UltimateArity^p}$    # Determine expected density if all data points would be randomly distributed
    $QFB \leftarrow 2 \cdot \left(\frac{ExpectedRandomDensity}{\frac{1}{n}\sum_{i=1}^{n} ads_i}\right) \cdot 100$    # Calculate QFB by taking the ratio of the expected and observed density
    **return** *QFB*
**end**

*Harmonic Mean Detection of HDAs (HMDH)*: The HMDH framework starts in a similar way as IPP, except it takes WeightCorrection as its single input parameter (QFB and QD are not relevant here). As in IPP, *aas* and *ads* are calculated using an underlying AD algorithm. Subsequently the vectors are reversed and rescaled, so that it can be expected that the most extreme HDAs will have a high score for both their $aas_g$ and $ads_g$. As a next step the optimal balance between these two vectors (representing individual anomalousness and density of the neighborhood) is determined. This is done by calculating for each case the harmonic mean between these two values, so as to obtain the individual HDA scores. The intuition behind this is that it will find an optimally balanced HDA score: a case having a high score for both $aas_g$ (i.e. it is anomalous) and $ads_g$ (i.e. it is located in a high-density area) will end up having a high HDA score. The harmonic mean ensures a low score if either of these values is low, so only HDAs will get assigned a high score. As a final step the HDA score vector is reversed so that, similar to IPP, the lowest scores represent the most anomalous occurrences.

The WeightCorrection offers three ways of calculating the harmonic mean. Option None simply calculates for each case an unweighted mean of $aas_g$ and $ads_g$. However, early experiments showed that if some classes (or combinations thereof in case of multiple categorical attributes) are unevenly distributed in the space while others are densely packed, the *ads* scores tend to have too much influence on the HDA scores. Option SSE therefore calculates the relative Shannon information entropy of the (combinations of the) classes, resulting in a value between 0 and 1 that is used as the weight for *ads* (with *aas* having a weight of 1). Option SDEN first uses the *aas* density score to calculate the arithmetic mean of the density per class (or class combination). Subsequently, the fraction of the highest average density and the harmonic mean of the remaining average densities yields a value between 0 and 1 that, similar to SSE, is used as the weight for *ads*.

IV. EVALUATION

This section presents the evaluation of the various frameworks and algorithms using the ROC/PRC AUC [35, 36, 37] and confusion matrix based metrics [1, 39]. Table I shows the characteristics of the datasets used. Synthetic sets had to be created for the current research, since no publicly available datasets could be found with explicitly labeled high-density anomalies. Anomalies were manually injected in such a way that they represent clear-cut HDAs. The Polis set is a real-world income dataset [17] without ground-truth and thus used here for an exploratory analysis. It contains 3 numerical attributes and 1 categorical attribute.

All experiments were conducted in R 3.6.1, RStudio 1.2.5001 and packages pROC 1.15.3, precrec 0.10.1, fmsb 0.6.3, dbscan 1.1-5, SECODA 0.5.4, DDoutlier 0.1.0 and spoutlier 1.0. Algorithms were generally executed with their standard settings. However, spoutlier's QSP was run using 3000 samples (instead of the standard 20) for more stable results, and SECODA was run without pruning to obtain the most precise scores. Unless stated otherwise, HMDH uses SECODA as its underlying algorithm.

TABLE I. OVERVIEW OF USED DATASETS

| Dataset | Nature | Data types | # Cases | # HDAs | Anomaly types (HDAs) |
|---|---|---|---|---|---|
| Gleuf | Simulated | 3 num, 1 categ | 25853 | 6 | VI |
| NoisyHelix | Simulated | 3 num, 1 categ | 9665 | 15 | VI |
| Multiset4D | Simulated | 3 num, 1 categ | 7853 | 22 | VI |
| Multiset5D | Simulated | 3 num, 2 categ | 70767 | 40 | II, V, VI |
| Polis dataset | Real-world | 3 num, 1 categ | 304726 | Unknown | (no explicit labels) |

NoisyHelix, Multiset4D and Multiset5D contain too many HDAs to be understandably plotted and discussed, necessitating the use of evaluation metrics to obtain a clear insight into the functional performance of the algorithms. Some idiosyncrasies of anomaly detection should be taken into account here. As a result of the extremely imbalanced class distribution (i.e. few anomalies, many normal cases), evaluation metrics such as ROC/PRC AUC, accuracy and specificity are often intrinsically high. Consequently, when studying these metrics it is important to also take the numbers to the right of the decimal point into account. Moreover, for AD it is crucial to analyze the *early retrieval* results, i.e. the validity of the most extreme anomaly scores [36], because these cases are typically classified as the anomalies that demand further action. The early retrieval performance is especially relevant in this study because the difference between regular AD algorithms and HDA frameworks are compared, both of which can be expected to detect HDAs – the main difference being the amount of false positives. The important early retrieval metrics are those that are not intrinsically high, e.g. F1, sensitivity and precision. The partial (left) area of the ROC curve is also relevant in this context, as this focuses on the most extreme scores and helps to analyze false positives [35, 37].

*Gleuf*: This set is shown in Fig. 3 and contains 6 high-density anomalies of Type VI (2 blue cases in the red cluster and 4 red cases in the blue cluster) as well as a large number of isolated cases that in the context of this study represent uninteresting noise. The IPP framework (regardless of it using KNN-AGG or SECODA) detects all 6 anomalies amongst its top 6 results (i.e. the 6 cases with the most extreme anomaly scores). The HMDH (SDEN) framework also performs well w.r.t. its early retrieval results, by detecting 3 true HDAs amongst its top 6 results and 5 amongst its top 10. However, regular (equiwidth) SECODA, which also targets isolated cases, does not detect any HDAs amongst its top 40 results, and detects only 1 HDA amongst the top 100 results. KNN-AGG detects 1 HDA amongst its top 40 results and 2 amongst its top 100 results. This relatively poor performance is the result of the many isolated noise cases present in the set, which are declared anomalous by regular AD algorithms such as KNN-AGG, QSP and SECODA, but not deemed relevant in the context of this study. Equidepth SECODA performs similar to HMDH SDEN, except the 6th HDA gets assigned an extremely non-anomalous score.

The impact of the presence of noise on detection performance is clearly illustrated in Fig. 3. For visualization purposes this plot discards one numerical dimension. The left sub-figure highlights all 6 true HDAs in light blue, which are correctly identified by IPP as the top 6 anomalies. The right sub-figure highlights the top 10 anomalies as returned by traditional AD approaches that view isolated case as anomalies (KNN-AGG and SECODA returned identical results here).

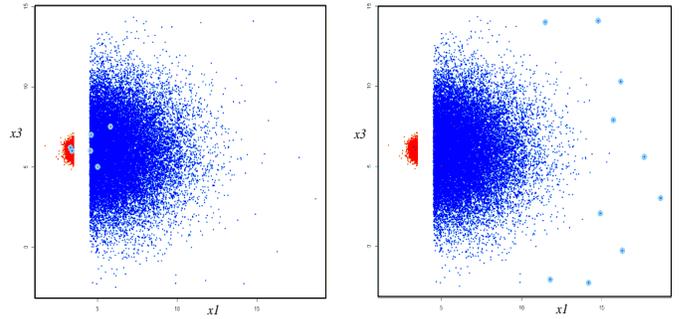

Fig. 3. Gleuf set (6 HDAs at the left and 10 noise cases at the right)

When the numerical $x2$ attribute is discarded from the analysis (as in Fig. 3), the first HDA is at the 11th position in terms of anomaly scores returned by regular SECODA. When all three numerical attributes are included in the analysis – and data points thus get scattered more – this effect is significantly more extreme, with the first HDA being positioned at the 41th position for SECODA. This clearly illustrates that a classic density or distance-based AD approach identifies the many isolated noise cases as anomalies and will thus not include the HDAs amongst the early retrieval results. It tangibly demonstrates why regular low-density or distance-oriented AD is not suited for detecting HDAs, especially when dealing with noisy or large datasets. Moreover, the effect is exacerbated if more attributes are added, as cases get dispersed more with increasing dimensionality.

*NoisyHelix*: This set features a helix pattern, a substantial amount of noise and HDAs in the 'wrong' cluster. Fig. 4 and 5 show a plot and the ROC curves respectively. Table A in the Appendix presents the (partial) ROC/PRC AUC values. Table B shows the early retrieval performance using various confusion matrix based metrics, which require a threshold on the scores to declare cases either normal or anomalous. Two thresholds are used, namely a cut-point based on the number of true HDAs and one on Youden's best ROC threshold [29]. Used metrics are sensitivity, specificity, precision, accuracy, F1 (harmonic mean of precision and sensitivity), Matthews correlation coefficient, Cohen's Kappa, GMRP (geometric mean of recall and precision) and HMFM (harmonic mean of four metrics, i.e. sensitivity, specificity, precision and accuracy) [35, 36, 37, 39].

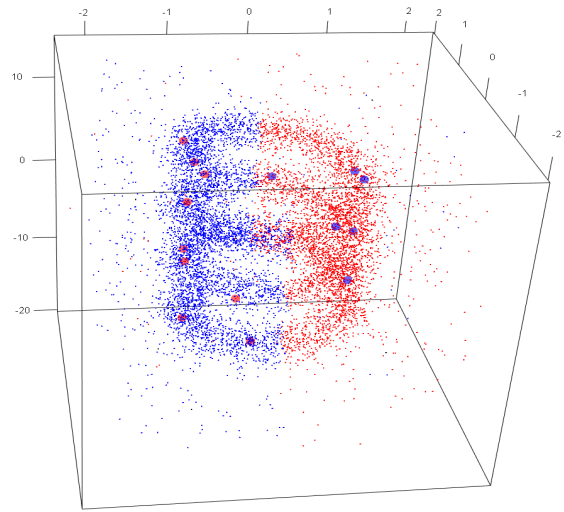

Fig. 4. Plot of NoisyHelix with HDAs (shown as large data points), seemingly located in a cluster of the wrong class (shown as color)

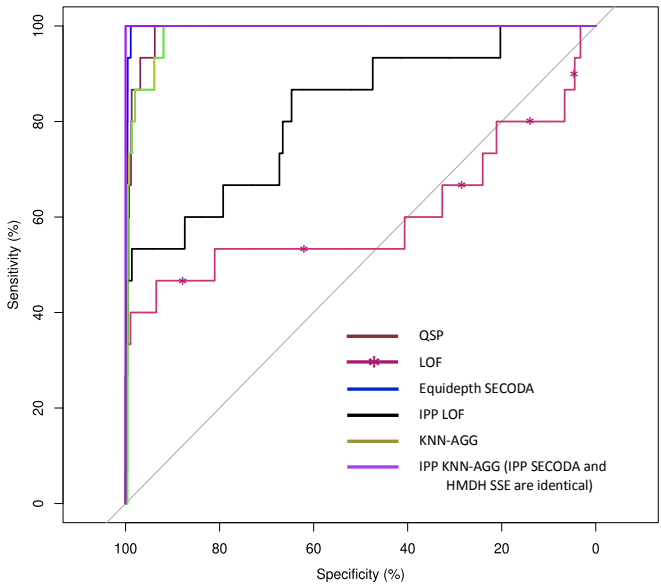

Fig. 5. ROC curves of NoisyHelix analysis

The AUC values indicate that the IPP and HMDH frameworks demonstrate the best overall performance, regardless of the underlying AD algorithm (except for LOF). Regular distance- and density-oriented AD performs relatively poorly, while equidepth SECODA yields medium performance. A more tangible insight is offered by the early retrieval results in Appendix Table B. These results also clearly illustrate that regular distance- and density-based algorithms perform poorly, with low values for all relevant metrics (note that those with high values will always be high in this context due to the way they are calculated). Table B shows the results for KNN-AGG and SECODA, but the other non-HDA algorithms demonstrated a similar poor performance. The precision, sensitivity (recall), F1 and various other measures of KNN-AGG using the true number of HDAs are all 0, meaning amongst the 15 cases with the most extreme anomaly scores 0 true HDAs are present. The sensitivity of SECODA is approximately 0.0667, which means that of the 15 true HDAs only 1 case is recognized as such. The precision of 0.0625 points to a very high degree of false positives.

For an analyst scrutinizing the most extreme scores, this means most of his or her time would be spent on the wrong cases. In this small dataset this could be overcome by simply spending more time, but in large sets with millions of data points the number of false positives would be so great as to make any manual analysis task practically infeasible. The HDA frameworks, which are designed specifically for this problem, perform very well in this regard. They exhibit high scores for all metrics, including those that are not naturally high in an AD context.

*Multiset4D*: This set consists of various clusters of a different class (in Fig. 6 represented by color). Fig. 7 shows the ROC curves. The AUC values of Table E in the Appendix indicate that the IPP framework yields the best overall performance, regardless of the underlying AD algorithm (again with the exception of LOF). This is followed closely by a good performance of the HMDH SDEN framework. However, HMDH SSE and HMDH None perform poorly. Distance- and density-oriented AD also perform less well (note that AUCs are often intrinsically high in AD due to the imbalanced distribution). Equidepth SECODA performs slightly better than the standard equiwidth variant.

Due to space limitations the confusion matrix based metrics could not be included in detail in the Appendix. However, the evaluations demonstrate similar results as for NoisyHelix and Multiset5D. Regular distance- and density-based algorithms perform poorly, with low values for all relevant metrics. To make tangible what this entails for KNN-AGG: of the 22 true HDAs only 2 are detected (yielding a sensitivity of 0.09). Of the 22 most extreme anomaly scores only 2 cases are true HDAs, with no less than 20 cases being false positives (yielding a precision of 0.09). In contrast, IPP with KNN-AGG obtained scores of 1, representing perfect results.

*Multiset5D*: Fig. 8 shows this is a dataset with very strict patterns and low levels of variation and fully random noise. The results in Appendix Table C make clear this is a challenging set for the frameworks to analyze, with IPP again demonstrating the best functional performance. For this set IPP is the only HDA framework performing better than the baseline algorithms. The three HMDH approaches and equidepth SECODA all perform very poorly. Somewhat surprisingly, the HMDH SDEN approach, which performed very well for the other sets, fails here and performs worse than the classic distance- and density-based approaches.

The AUC scores of the HMDH SSE and weightless HMDH None approaches are both approximately 60%, meaning that the Shannon entropy does not provide much value in adjusting the weights for this set. The non-HDA algorithms perform relatively well in terms of AUCs, which can be attributed to the fact that the set does not contain many noise cases that mask the true HDAs. However, the results in Table D show that the early retrieval performance of these algorithms is still poor. Although it performs better than regular LOF, IPP LOF performs relatively poor when compared to the other IPP instantiations.

As an anecdote, the IPP framework also proved its value by discovering two errors in this dataset. During the preparation phase these cases were erroneously labeled by the author as HDAs, but in reality were isolated cases. When IPP seemingly could not detect these 'HDAs', closer inspection revealed that these data points were in fact isolated and had faulty labels, after which this could be corrected.

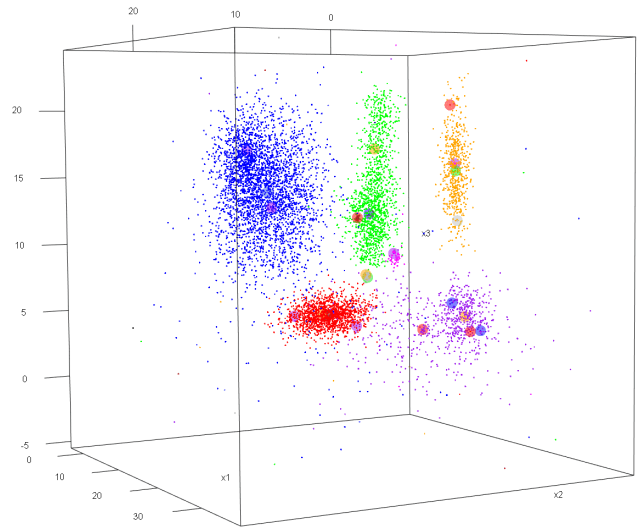

Fig. 6. Various Multiset4D clusters containing high-density Type VI anomalies (shown as large data points correctly detected by IPP with KNN-AGG) that are of a class (color) from a different cluster

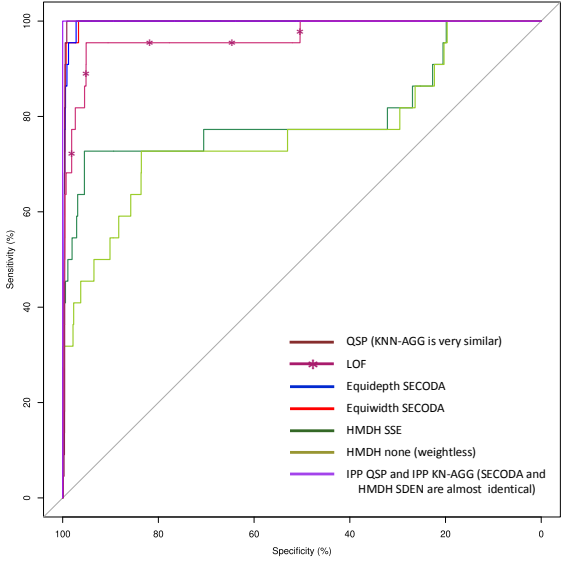

Fig. 7. ROC curves of Multiset4D analysis

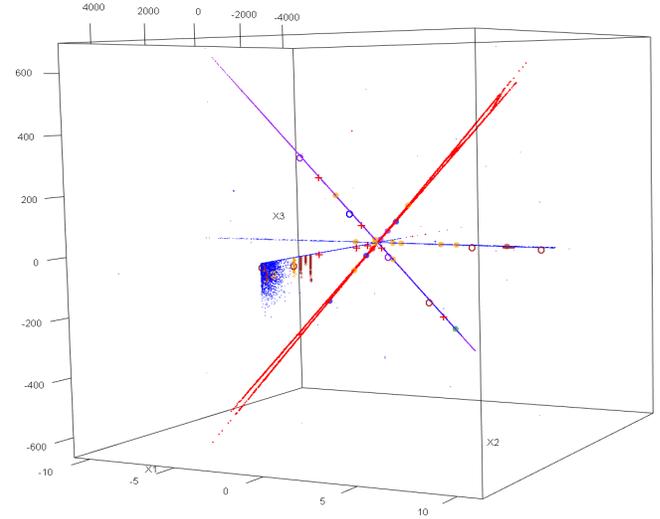

Fig. 8. Plot of Multiset5D with anomalies shown as large symbols (with the color and symbols denoting the different qualitative classes)

*Polis dataset*: This is a real-world income dataset with 1 social security code (qualitative), 1 wage and 2 social charge (all quantitative) attributes. The set demonstrates strong patterns based on governmental laws and regulations. No ground-truth is available, but the exploratory analysis clearly demonstrates the differences between the traditional AD detection and HDA detection frameworks. Fig. 9 shows the plots of regular and IPP anomaly detection (the bottom part zooms in on the high-density cloud in the center). The left plot illustrates the detection results of regular SECODA, which, as can be seen, detects mainly isolated cases. The right plot shows the detection results of IPP with SECODA, which mainly detects anomalies within the high-density center. A similar effect can be seen when conducting this analysis with e.g. KNN-AGG or QSP. The different versions of the HMDH framework yield meaningful results as well, albeit somewhat less diverse than IPP (i.e. many anomalies identified by HMDH are positioned relatively close to each other).

Summarizing the results, the IPP framework consistently outperforms the other approaches. The HMDH framework performed well on most sets, with the SDEN weight adjustment method consistently performing best across the HMDH versions. However, HMDH performed poorly on Multiset5D because none of the weight adjustment methods worked properly for this set. The traditional, general-purpose AD algorithms were often also able to identify the HDAs. However, in large sets they tend to yield too many false positives to be effectively used, as is demonstrated by the early retrieval results that then predominantly contain isolated cases. Of these non-HDA algorithms, distance and global density based approaches perform relatively well (albeit never as good as when used within IPP), while the LOF algorithm did not deliver good results. Finally, SECODA with equidepth discretization performed better than equiwidth discretization in most situations, but demonstrated inconsistent performance across datasets and despite good early retrieval results sometimes had difficulty in detecting the last, more nuanced HDA cases.

Note that some of the results can be improved further. QSP and SECODA are optimized for scalability at the cost of precision. Increasing the number of samples for QSP and the number of iterations for SECODA makes the results more precise, bringing their performance closer to that of KNN-AGG. This holds both for their use as independent algorithms and within the IPP and HMDH frameworks. If the underlying analysis results get more precise, the results of a framework that leverages them will also get more precise.

## V. DISCUSSION

High-density anomalies have an understandable and meaningful interpretation: they represent deviant occurrences that hide in normality. In some respects they are anomalous, in other respects they belong to the most normal of cases. As such, a HDA can be seen as an occurrence that tries to mask its deviating nature by exhibiting the most normal behavior possible for that situation. HDAs are unusual cases that nonetheless may easily go unnoticed because they 'hide in numbers', namely in high-density areas. Contrary to traditional anomalies, which are typically conceived as isolated, low-density occurrences, HDAs are positioned amongst the most normal data points.

Note that "most normal" and "high-density" are somewhat relative concepts. In a given dataset the highest density regions may contain no HDAs at all as these may all be positioned in areas that are only moderately dense. Also, a moderately dense neighborhood may contain an extreme anomaly (e.g. a unique class) and a highly dense region a modest anomaly (e.g. a rare class). The HDA detection task is in essence a trade-off and balancing problem between case anomalousness and neighborhood density, with the most normal (high-density) regions not necessarily hosting HDAs. The IPP and HMDH frameworks each have their own way of managing this balancing task, with the QuantileFilterBoost and weight adjustments respectively playing important roles.

As discussed in the Introduction, HDA detection can typically be valuable for misbehavior detection and data quality use cases, where the focal deviants are likely to also demonstrate very normal behavior. In addition, identification of HDAs may be useful when the dataset contains a high degree of noise that one is not interested in – as clearly illustrated by the Gleuf example of Fig. 3. Existing algorithms for anomaly detection, such as distance- and density-based approaches, will generally declare the isolated cases anomalous. However, these may thus represent uninteresting statistical noise or simple extreme cases in which one takes no interest. Especially in analyses of large datasets it will be necessary to then actively avoid wasting time on these noise cases. For example, governmental data registers, such as the Polis administration, often include the whole population and can therefore be expected to contain considerable statistical variation, even though many of the isolated cases will probably not be erroneous. HDA detection can help to focus on deviations amongst the most normal cases and to avoid spending valuable time on these univariately or multivariately isolated Type I, III and IV outliers. Without such methods it will otherwise be very difficult to detect high-density anomalies in large databases, because the noisy data points will mask the true HDAs. The focal analysis results (i.e. the early retrieval results that one will typically study in detail) will consequently contain a large number of false positives and miss true HDAs. In such situations an approach directly targeting HDAs is a downright necessity in order to conduct an effective and efficient analysis.

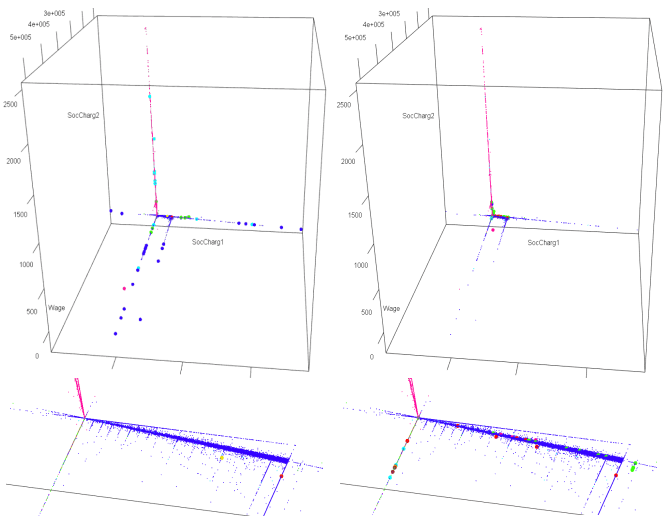

Fig. 9. Real-life Polis income dataset. Large data points represent the top 40 anomalies as detected by SECODA (left) and IPP SECODA (right). Anomalies detected by IPP are positioned in the high-density area.

Having said this, an interesting property of the IPP algorithm is that it still returns the isolated cases. The HDAs get assigned the lowest scores and can thus easily be identified by the analyst. However, the algorithm operates in such a way that the cases with the highest scores prove to be the isolated cases. The last statements of the pseudocode, just before returning *hds*, add the filtered-out isolated cases to *hds* with high values. Both the low and high ends of the score vector thus contain anomalies.

LOF did not perform very well in the context of HDA analysis, neither as a stand-alone technique nor as an underlying algorithm of a HDA framework. Many HDAs are not assigned proper anomaly scores by LOF and only get declared anomalous using a high threshold, resulting in a relatively large number of false positives. An important reason for this is that LOF does not yield a global outlier assessment for the individual cases (yielding a suboptimal *aas*). This is due to its local orientation, in which an individual data point that is positioned close to a given cluster may still get assigned a score that implies it is isolated (which it only is from a local perspective). Perhaps this may be relevant to some HDA situations, but the HDAs in this study's datasets are mainly anomalous from the perspective of the global numerical space. Because of its local focus LOF also does not effectively detect globally isolated noise cases (yielding a suboptimal *ads*), which means these cannot be filtered away effectively. For example, many of the noise cases scattered throughout the space of the NoisyHelix set are not acknowledged as outliers and therefore cannot effectively be removed. In short, a HDA framework works well if its underlying algorithm suits the set and the problem at hand, because both *aas* and *ads* need to be correct.

In this study all the dataset's numerical attributes are used to determine the neighborhood density, whereas the full set of attributes (including the categorical ones) are used to verify whether the case is anomalous. This analysis can be done unsupervised and without manually setting any input parameters, which is a favorable characteristic for a data mining method [22]. However, conceptually there is no reason why the neighborhood density needs to be assessed only in the numerical space, or why all attributes should be involved in assessing anomalousness. The definition of a high-density neighborhood could therefore be extended to take only a subset of the numerical variables into account and even to include categorical attributes. Likewise, the assessment of how anomalous a case is can be based on a flexible subspace. Note, however, that this requires explicit input from and domain knowledge of the data analyst, because the dentributes (density attributes) in such a scenario are not necessarily numerical and thus can no longer be automatically determined. Extended versions of the algorithmic frameworks then thus require the user to explicitly parameterize which attributes should be used for the high-density assessment (normalness of the neighborhood) and which ones should be used for the low-density assessment (anomalousness of the individual cases).

Finally, from a technical point of view the IPP framework allows being executed in a distributed fashion. For instance, given the standard value for the QuantileDenominator setting the algorithm will perform 100 iterations. These can largely be done independently, e.g. on 100 machines.

## VI. Conclusion

As a first contribution, this study has explored a novel perspective on anomalies, which differs from the traditional view of anomalies as isolated, low-density data points. A HDA is an anomaly in a highly normal neighborhood, the detection of which requires an approach to balance the degree of deviation and neighborhood density. The concept of HDAs is not only new from a theoretical perspective, but also relevant for practical use cases. A second contribution of this research is the introduction of two algorithmic frameworks (IPP and HMDH) and several variations thereof, which are designed to identify HDAs. A third contribution is the evaluation of these frameworks and of several existing algorithms used as a baseline. A final contribution is the creation of four publicly available simulated sets that can be used for future research (see Remarks for hyperlink).

There are several topics for future research. Firstly, the HMDH framework can be improved. The experiments demonstrate they offer good detection performance on various sets. However, the performance is unreliable, and alternative or improved methods for determining the weights should be studied.

Secondly, experiments with the IPP and HMDH frameworks can be conducted on high-dimensional datasets and in combination with additional general-purpose AD algorithms, both as underlying algorithms for the frameworks and as baseline algorithms for comparison of results.

A third topic for future research is setting a proper threshold on the gradual scores to explicitly declare cases a high-density anomaly This is especially important in this context. The experiments show that if the HDA algorithm works properly, the cases with the most extreme scores are indeed all HDAs. However, at some position in the resulting score vector all true HDAs will have been declared anomalous and the less extreme scores will be false positives if the threshold is set too high. Obviously, threshold setting is an issue for any algorithm returning a continuous anomaly score vector, but the problem is more prevalent in the HDA setting. With regular distance- or density-based algorithms the cases with less extreme scores will be increasingly normal and less isolated – indeed as a true continuous phenomenon. However, with HDA analysis false positives will suddenly manifest themselves and perfectly normal cases will be declared anomalous. Future research should therefore aim to find approaches to set a proper threshold.

Another topic to study is the way how aggregate anomalies (groups or collectives) can manifest themselves as HDAs. These aggregates, represented in the bottom row of Fig. 2 and discussed in [38], have been ignored in this research. A final topic for future research concerns extending the frameworks with manually setting the subspaces as discussed in Section V.

REMARKS AND ACKNOWLEDGMENTS

The web link to the labeled datasets and R code to analyze them are available for download from the *SECODA resources for R* section at www.foorthuis.nl, and includes SECOHDA implementations of the HDA frameworks. This research has been supported by HEINEKEN, UWV and Loonaangifteketen.

APPENDIX

The following tables provide additional details and reproducibility information.

| | Algorithm | | | | | | | | | | | |
|---|---|---|---|---|---|---|---|---|---|---|---|---|
| | HMDH SDEN | HMDH SSE | HMDH None | IPP SECODA | SECODA | SECODA ED | IPP QSP | QSP | IPP KNN-AGG | KNN-AGG | IPP LOF | LOF |
| ROC AUC | 99.9896373% | 100% | 99.9972366% | 100% | 99.1347150% | 99.6777202% | 99.9744387% | 98.8307426% | 100% | 98.4960276% | 82.0725389% | 60.4069084% |
| ROC partial AUC for 100-90% specificity | 99.9454595% | 100% | 99.9854559% | 100% | 95.4458686% | 98.3037906% | 99.8654668% | 93.8460140% | 100% | 92.0843560% | 74.8095628% | 69.2246159% |
| PRC AUC | 99.9999839% | 100% | 99.9999957% | 100% | 99.9986303% | 99.9994983% | 99.9999603% | 99.9981542% | 100% | 99.9976034% | 99.9589285% | 99.8496390% |

Table A. AUC values for NoisyHelix analysis.

| | | IPP SECODA, IPP KNN-AGG & HMDH SSE | |
|---|---|---|---|
| | | Threshold based on # of true HDAs | Threshold based on best Youden ROC |
| Metric | Sensitivity/Recall | 1 | 1 |
| | Specificity | 1 | 1 |
| | Precision/PPV | 1 | 1 |
| | Accuracy | 1 | 1 |
| | F1 measure | 1 | 1 |
| | Matthews CC | 1 | 1 |
| | Cohen's Kappa | 1 | 1 |
| | GMRP | 1 | 1 |
| | HMFM | 1 | 1 |

| | | HMDH SDEN | |
|---|---|---|---|
| | | Threshold based on # of true HDAs | Threshold based on best Youden ROC |
| Metric | Sensitivity/Recall | 0.933333333 | 1 |
| | Specificity | 0.999896373 | 0.998756477 |
| | Precision/PPV | 0.933333333 | 0.555555556 |
| | Accuracy | 0.999793068 | 0.998758407 |
| | F1 measure | 0.933333333 | 0.714285714 |
| | Matthews CC | 0.933229706 | 0.744892415 |
| | Cohen's Kappa | 0.933229706 | 0.713714455 |
| | GMRP | 0.933333333 | 0.745355992 |
| | HMFM | 0.965444857 | 0.832901577 |

| | | SECODA ED | |
|---|---|---|---|
| | | Threshold based on # of true HDAs | Threshold based on best Youden ROC |
| Metric | Sensitivity/Recall | 0.266666667 | 1 |
| | Specificity | 0.998860104 | 0.989015544 |
| | Precision/PPV | 0.266666667 | 0.123966942 |
| | Accuracy | 0.997723745 | 0.989032592 |
| | F1 measure | 0.266666667 | 0.220588235 |
| | Matthews CC | 0.265526770 | 0.350150300 |
| | Cohen's Kappa | 0.265526770 | 0.218429826 |
| | GMRP | 0.266666667 | 0.352089395 |
| | HMFM | 0.420900990 | 0.360722313 |

| | | KNN-AGG | |
|---|---|---|---|
| | | Threshold based on # of true HDAs | Threshold based on best Youden ROC |
| Metric | Sensitivity/Recall | 0 | 1 |
| | Specificity | 0.998445596 | 0.919170984 |
| | Precision/PPV | 0 | 0.018867925 |
| | Accuracy | 0.996896017 | 0.919296430 |
| | F1 measure | 0 | 0.037037037 |
| | Matthews CC | -0.001554404 | 0.131692250 |
| | Cohen's Kappa | -0.001554404 | 0.034094403 |
| | GMRP | 0 | 0.137360564 |
| | HMFM | 0 | 0.071205133 |

| | | SECODA | |
|---|---|---|---|
| | | Threshold based on # of true HDAs | Threshold based on best Youden ROC |
| Metric | Sensitivity/Recall | 0.066666667 | 1 |
| | Specificity | 0.998445596 | 0.934715026 |
| | Precision/PPV | 0.0625 | 0.023255814 |
| | Accuracy | 0.996999483 | 0.934816348 |
| | F1 measure | 0.064516129 | 0.045454545 |
| | Matthews CC | 0.063047944 | 0.147436626 |
| | Cohen's Kappa | 0.063015027 | 0.042550180 |
| | GMRP | 0.064549722 | 0.152498570 |
| | HMFM | 0.121195351 | 0.086693476 |

| | | HMDH None | |
|---|---|---|---|
| | | Threshold based on # of true HDAs | Threshold based on best Youden ROC |
| Metric | Sensitivity/Recall | 0.933333333 | 1 |
| | Specificity | 0.999896373 | 0.999689119 |
| | Precision/PPV | 0.933333333 | 0.833333333 |
| | Accuracy | 0.999793068 | 0.999689602 |
| | F1 measure | 0.933333333 | 0.909090909 |
| | Matthews CC | 0.933229706 | 0.912729021 |
| | Cohen's Kappa | 0.933229706 | 0.908936732 |
| | GMRP | 0.933333333 | 0.912870929 |
| | HMFM | 0.965444857 | 0.952240050 |

Table B. Early retrieval results for NoisyHelix analysis.

| | Algorithm | | | | | | | | | | | |
|---|---|---|---|---|---|---|---|---|---|---|---|---|
| | HMDH SDEN | HMDH SSE | HMDH None | IPP SECODA | SECODA | SECODA ED | IPP QSP | QSP | IPP KNN-AGG | KNN-AGG | IPP LOF | LOF |
| ROC AUC | 88.8580387% | 61.6405687% | 58.2851316% | 99.9977731% | 99.9044212% | 89.7699782% | 99.9279271% | 99.8278239% | 99.9999293% | 99.9203982% | 99.1906556% | 98.4309740% |
| ROC partial AUC for 100-90% specificity | 75.5825215% | NA | NA | 99.9882796% | 94.4969538% | 91.7825806% | 99.6206689% | 99.0938099% | 99.9996279% | 99.5810429% | 97.3215395% | 97.0495523% |
| PRC AUC | 99.9923298% | 99.9706685% | 99.9679945% | 99.9999987% | 99.9999459% | 99.9902886% | 99.9999591% | 99.9999024% | 99.9999999% | 99.9999550% | 99.9994997% | 99.9989423% |

Table C. AUC values for Multiset5D analysis.

| | | IPP SECODA | |
|---|---|---|---|
| | | Threshold based on # of true HDAs | Threshold based on best Youden ROC |
| Metric | Sensitivity/Recall | 0.95 | 1 |
| | Specificity | 0.999971722 | 0.999208223 |
| | Precision/PPV | 0.95 | 0.416666667 |
| | Accuracy | 0.999943476 | 0.999208671 |
| | F1 measure | 0.95 | 0.588235294 |
| | Matthews CC | 0.949971722 | 0.645241629 |
| | Cohen's Kappa | 0.949971722 | 0.587906452 |
| | GMRP | 0.95 | 0.645497224 |
| | HMFM | 0.974338847 | 0.740523471 |

| | | HMDH SDEN | |
|---|---|---|---|
| | | Threshold based on # of true HDAs | Threshold based on best Youden ROC |
| Metric | Sensitivity/Recall | 0.075 | 0.75 |
| | Specificity | 0.998727502 | 0.869088184 |
| | Precision/PPV | 0.032258065 | 0.003229626 |
| | Accuracy | 0.998205378 | 0.869020871 |
| | F1 measure | 0.045112782 | 0.006431558 |
| | Matthews CC | 0.048370527 | 0.043574336 |
| | Cohen's Kappa | 0.044357366 | 0.005311913 |
| | GMRP | 0.049186938 | 0.049216053 |
| | HMFM | 0.086325212 | 0.012768619 |

| | | SECODA ED | |
|---|---|---|---|
| | | Threshold based on # of true HDAs | Threshold based on best Youden ROC |
| Metric | Sensitivity/Recall | 0.475 | 0.85 |
| | Specificity | 0.999703084 | 0.995093811 |
| | Precision/PPV | 0.475 | 0.089238845 |
| | Accuracy | 0.999406503 | 0.995011799 |
| | F1 measure | 0.475 | 0.161520190 |
| | Matthews CC | 0.474703084 | 0.274487079 |
| | Cohen's Kappa | 0.474703084 | 0.160661494 |
| | GMRP | 0.475 | 0.275414267 |
| | HMFM | 0.643975423 | 0.277926458 |

| | | KNN-AGG | |
|---|---|---|---|
| | | Threshold based on # of true HDAs | Threshold based on best Youden ROC |
| Metric | Sensitivity/Recall | 0.25 | 1 |
| | Specificity | 0.999575834 | 0.997539836 |
| | Precision/PPV | 0.25 | 0.186915888 |
| | Accuracy | 0.999152147 | 0.997541227 |
| | F1 measure | 0.25 | 0.314960630 |
| | Matthews CC | 0.249575834 | 0.431805563 |
| | Cohen's Kappa | 0.249575834 | 0.314307547 |
| | GMRP | 0.25 | 0.432337701 |
| | HMFM | 0.399949090 | 0.478759187 |

| | | SECODA | |
|---|---|---|---|
| | | Threshold based on # of true HDAs | Threshold based on best Youden ROC |
| Metric | Sensitivity/Recall | 0.2 | 1 |
| | Specificity | 0.999547556 | 0.993156786 |
| | Precision/PPV | 0.2 | 0.076335878 |
| | Accuracy | 0.999095624 | 0.993160654 |
| | F1 measure | 0.2 | 0.141843972 |
| | Matthews CC | 0.349632239 | 0.085550525 |
| | Cohen's Kappa | 0.349632239 | 0.014531431 |
| | GMRP | 0.2 | 0.276289482 |
| | HMFM | 0.333295620 | 0.248234790 |

| | | IPP KNN-AGG | |
|---|---|---|---|
| | | Threshold based on # of true HDAs | Threshold based on best Youden ROC |
| Metric | Sensitivity/Recall | 0.975 | 1 |
| | Specificity | 0.999985861 | 0.999985861 |
| | Precision/PPV | 0.975 | 0.975609756 |
| | Accuracy | 0.999971738 | 0.999985869 |
| | F1 measure | 0.975 | 0.987654321 |
| | Matthews CC | 0.974985861 | 0.987722614 |
| | Cohen's Kappa | 0.974985861 | 0.987647253 |
| | GMRP | 0.975 | 0.987729597 |
| | HMFM | 0.987331439 | 0.993781840 |

Table D. Early retrieval results for Multiset5D analysis.

| | Algorithm | | | | | | | | | | | |
|---|---|---|---|---|---|---|---|---|---|---|---|---|
| | HMDH SDEN | HMDH SSE | HMDH None | IPP SECODA | SECODA | SECODA ED | IPP QSP | QSP | IPP KNN-AGG | KNN-AGG | IPP LOF | LOF |
| ROC AUC | 99.9665173% | 80.6056349% | 76.7027316% | 99.9970978% | 99.5217144% | 99.5397082% | 100% | 99.5669890% | 100% | 99.5298406% | 96.4035709% | 96.6165937% |
| ROC partial AUC for 100-90% specificity | 99.9816702% | 81.0237683% | 70.0979850% | 99.9847252% | 97.4827074% | 97.5774113% | 100% | 97.7209948% | 100% | 97.5254769% | 97.2318946% | 91.6703275% |
| PRC AUC | 99.9999902% | 99.9020880% | 99.8877567% | 99.9999918% | 99.9986494% | 99.9987009% | 100% | 99.9986779% | 100% | 99.9986788% | 99.9807285% | 99.9880271% |

Table E. AUC values for Multiset4D analysis.